\documentclass{article}

\usepackage[main, final]{neurips_2025}

\usepackage[utf8]{inputenc} 
\usepackage[T1]{fontenc}    
\usepackage{hyperref}       
\usepackage{url}            
\usepackage{booktabs}       
\usepackage{amsfonts}       
\usepackage{nicefrac}       
\usepackage{microtype}      
\usepackage{xcolor}         
\usepackage{graphicx}
\usepackage{enumitem}
\usepackage{amsmath}
\usepackage{makecell}
\usepackage{colortbl}
\usepackage{adjustbox}
\usepackage{multirow}
\usepackage{pifont}
\usepackage[most]{tcolorbox}
\usepackage{fvextra}
\usepackage{amssymb}
\usepackage{hyperref}
\usepackage{fontawesome}

\definecolor{codegray}{gray}{0.95}

\definecolor{baselinecolor}{gray}{.9}

\title{RLFactory: A Plug-and-Play Reinforcement Learning Post-Training Framework for LLM Multi-Turn Tool-Use}

\author{
  RLFactory Team \\
  \\
  \href{https://github.com/Simple-Efficient/RL-Factory}{\faGithub\ \texttt{https://github.com/Simple-Efficient/RL-Factory}}
}

\begin{document}

\maketitle

\begin{abstract}

Large Language models (LLMs) exhibit strong advantages in basic reasoning, but face limitations in tasks that require interaction with the external environment. The "model-tool" collaboration paradigm has become a key solution by allowing models to learn to call external tools to obtain environmental feedback to improve task performance, with multi-round tool calls being a typical form of such tasks. However, reinforcement learning (RL) in multi-round tool call tasks faces challenges in tool call stability/adaptability (due to tool heterogeneity and interface issues) and reward computation diversity (due to different evaluation task requirements). To help the industry better study the tool call problem of LLMs, we propose RLFactory, a plug-and-play reinforcement learning post-training framework tailored for LLMs' multi-round tool calls. Its core design includes: (1) an asynchronous tool call mechanism based on asyncio, which greatly improves time efficiency; (2) a decoupled architecture that separates tool call and training modules to reduce environment setup costs; and (3) a diverse reward computation framework that supports rule-based, model judgment, and tool verification. RLFactory reconstructs the Markov Decision Process (MDP) state space by introducing observation markers (derived from tool feedback), achieving a closed-loop interaction between the model, tool, and environment. It implements a multi-round interactive workflow of "generate-parse-invoke-update" to ensure dynamic policy optimization. In a multi-line experiment on the Search-R1 project, RLFactory, using Qwen3-4B as the base model, achieved a test score of 0.486 on the NQ dataset, outperforming larger parameter models trained using the same techniques (e.g., Qwen2.5-7B-Instruct-GRPO, which scored 0.473). Furthermore, RLFactory increased training throughput by 6.8 times, demonstrating exceptional efficiency and stability. RLFactory provides a low-barrier, highly adaptable training framework for enhancing the multi-round tool usage of LLMs in real-world scenarios. The code is available on \href{https://github.com/Simple-Efficient/RL-Factory}{RL-Factory}.

\end{abstract}

\section{Introduction}
\label{sec:intro}
With the rapid iteration of large language models, their capabilities in natural language understanding, generation, and basic reasoning have been widely validated. However, when confronted with tasks that require up-to-date information retrieval, complex mathematical operations\cite{GPT5,mai2025agentrlscalinglaw}, or multi-step coordination\cite{dong2025agenticreinforcedpolicyoptimization,singh2025agenticreasoningtoolintegration}, relying solely on the knowledge embedded in the model's parameters often fails to meet the precision and reliability requirements. Against this backdrop, the "model-tool" collaborative paradigm has gradually become the core path to overcoming these limitations—by invoking external tools (such as search engines\cite{li2025searcho1agenticsearchenhancedlarge,jin2025search,wu2025mmsearchr1incentivizinglmmssearch}, code interpreters, vector database interfaces, etc.), models can expand their knowledge boundaries and enhance their task processing capabilities, thereby achieving better performance in practical application scenarios.

\begin{figure}[h]
    \centering
    \includegraphics[width=1\linewidth]{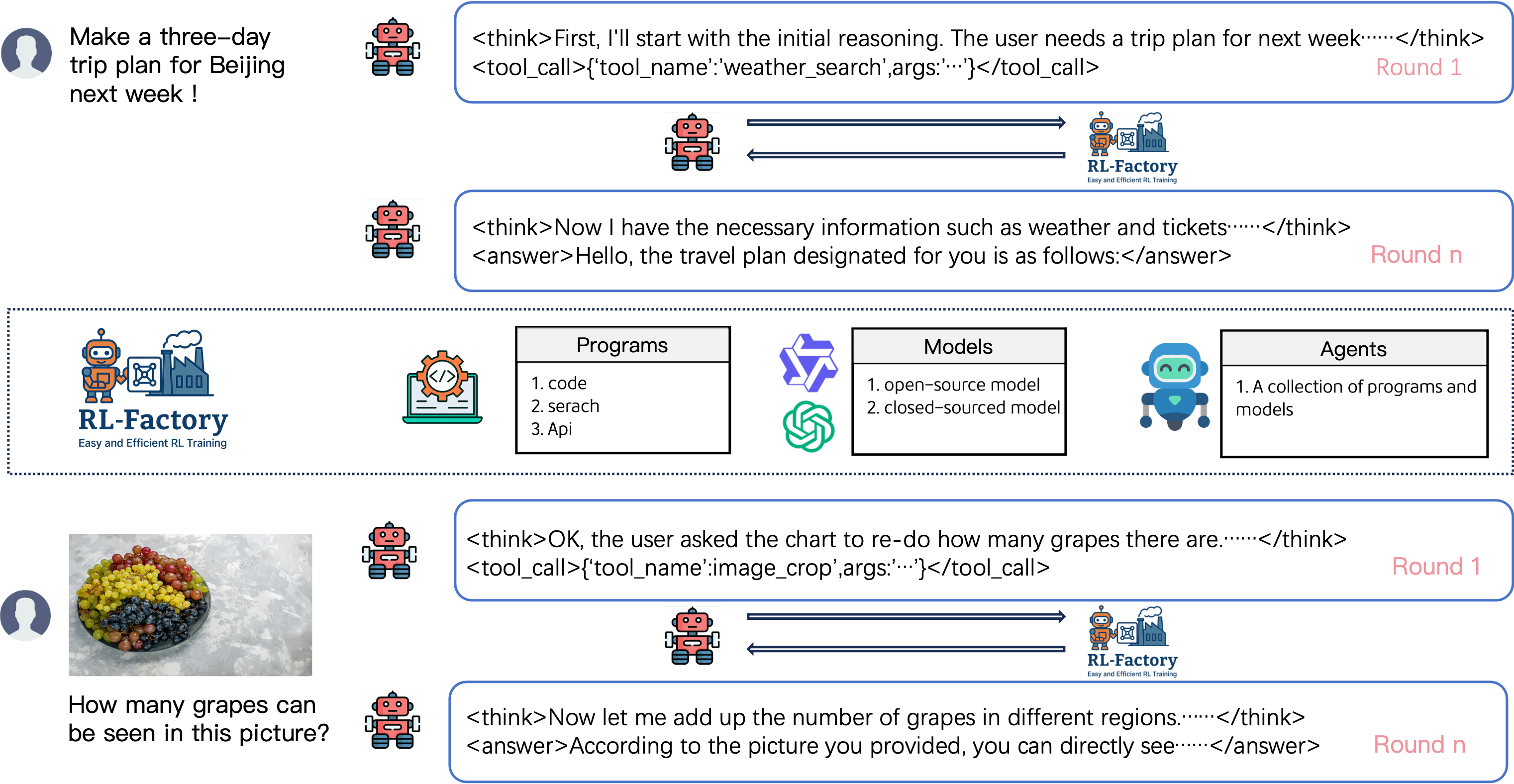}
    \vspace{-6mm}
    \caption{The "model-tool" collaborative paradigm.}
    \label{rlf_demo}
    \vspace{-10pt}
\end{figure}

In the "model-tool" collaborative paradigm, models need to dynamically adjust their invocation strategies based on the results returned by tools, often requiring multiple rounds of interaction to optimize outputs incrementally, as shown in Figure~\ref{rlf_demo}, making travel plans for users may require multiple calls and multiple tools. This makes multi-turn tool usage a typical form of such tasks. For example, deep search~\cite{zheng2025deepresearcherscalingdeepresearch,jin2025search} and travel planning~\cite{gundawar2024robust,chaudhuri2025tripcraft} are tasks that fall under the category of multi-turn tool usage. Meanwhile, large language model agents (LLM Agents) have evolved from merely invoking single tools to using LLMs as tools and constructing agentic graphs by invoking other agents. In this context, there is a pressing need in academia and industry for reinforcement learning post-training (RL Post-training) frameworks with strong targeted and plug-and-play characteristics to enhance model performance in the aforementioned vertical tasks. Additionally, recent models like Qwen3 have shown excellent performance in tool invocation and instruction compliance, making it feasible to bypass the SFT stage and directly conduct reinforcement learning training in vertical domain contexts.

Despite the clear research value and application prospects of multi-turn tool usage in reinforcement learning training, there remain numerous technical obstacles during practical implementation. Among them, challenges related to the stability and adaptability of tool invocation are particularly prominent: the framework-supported tool types encompass programs, models, and agents, with significant heterogeneity in interface specifications, response formats, and operational stability across different tools. Moreover, collaborative interactions between tools during multi-turn invocations are prone to interface issues. Concurrently, reward computation, as a core segment of reinforcement learning training, faces adaptation challenges in multi-turn tool usage scenarios—different tasks have varied logical demands for reward evaluation, with some relying on regulated computations from tool results and others requiring subjective judgments through model reasoning; a single reward computation approach is difficult to satisfy the diverse training needs.

To effectively address the core challenges at the tool invocation level, this framework has undergone targeted design in technical architecture and invocation mechanisms, yielding innovative solutions across multiple dimensions:

1. Asynchronous Tool Invocation Mechanism Design: We use the MCP mechanism to achieve unified registration management and design an asynchronous tool calling solution: the asynchronous parallel tool calling mechanism based on asyncio allows the model to initiate call requests to other tools while waiting for a tool to return results, significantly improving the time efficiency of tool calling. 

2. Decoupling of Tool Environment and Training System Architecture: By implementing a "complete decoupling" architecture design, the tool invocation module is separated from the training module into independent functional components, thereby substantially lowering the costs of tool environment setup.

3. Diverse Reward Computation Support Framework: This framework is constructed with diverse reward computation support schemes to address reward evaluation needs across different task contexts, offering both rule-based reward computation and model-based judgment reward computation.

\vspace{-5pt}

\section{Method}
\label{sec:method}

\subsection{Overall Design Concept}
RLFactory focuses on the post-training scenario of reinforcement learning (RL) for multiple rounds of tool invocation with large language models (LLMs). It aims to overcome the limitations and complexities in tool interaction adaptation, reward mechanism design, and development barriers associated with traditional frameworks. The core concept is to construct a layered modular architecture that decouples the training of tool invocation capabilities from diversified reward computation, thereby achieving a low-barrier "plug-and-play" development experience. By integrating asynchronous parallel interaction, flexible reward strategies, and a decoupled tool environment, the model autonomously learns the optimal strategy for multiple rounds of tool invocation, adapting to real-world scenarios such as deep search and complex task planning.

\begin{figure}[h]
    \centering
    \includegraphics[width=1\linewidth]{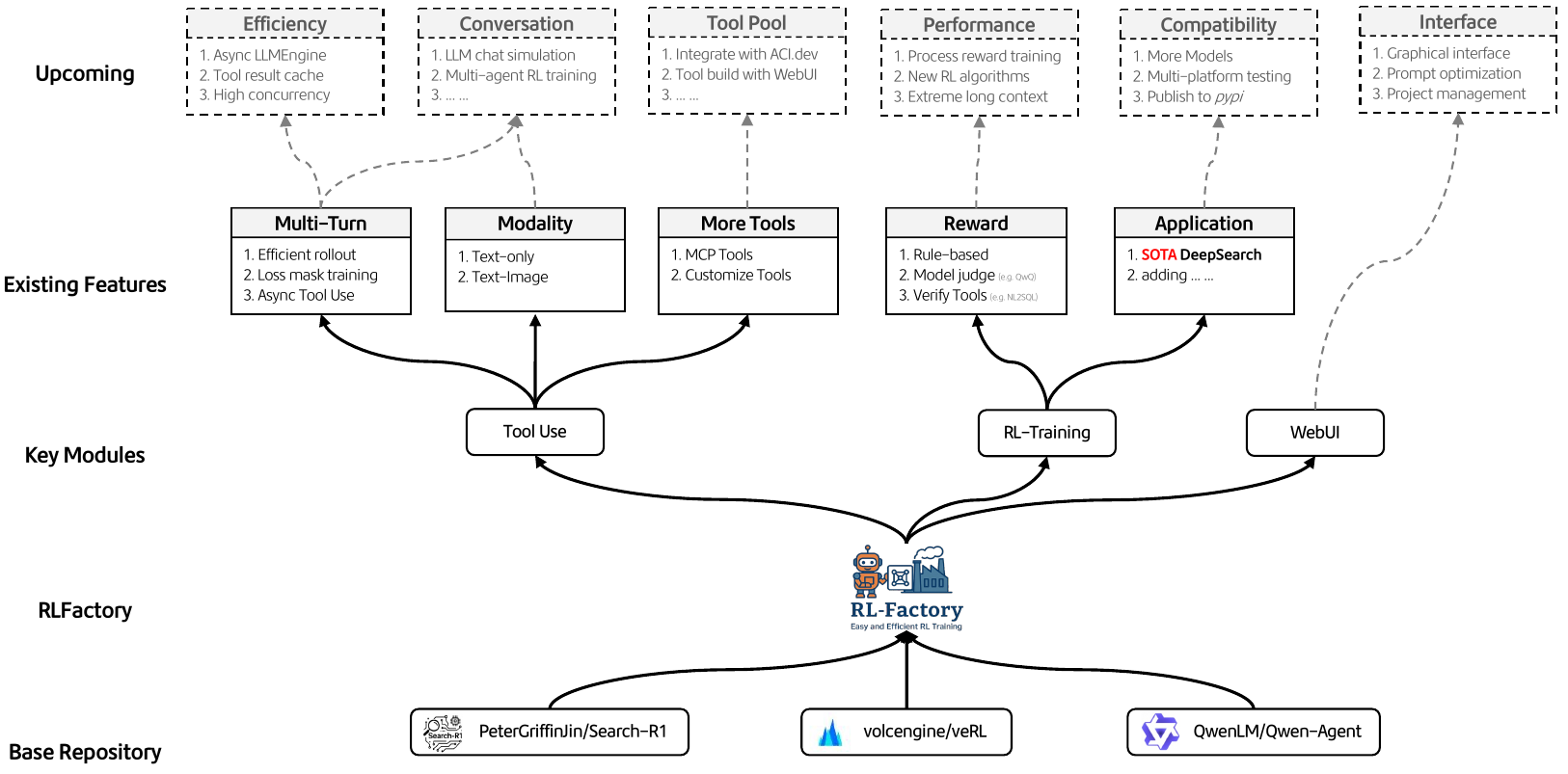}
    \vspace{-6mm}
    \caption{The framework of RLFactory.}
    \label{rlf_framework}
    \vspace{-10pt}
\end{figure}

As illustrated in Figure~\ref{rlf_framework}, the RLFactory is grounded in the Search-R1 tool invocation process, based on the veRL~\cite{sheng2024hybridflow} reinforcement learning training framework. It utilizes Qwen-Agent for tool construction and employs MCP for the unified registration management of tools. Considering the efficiency of reinforcement learning in multiple rounds of tool invocation, RLFactory implements three key modules: Tool Use, RL-Training, and WebUI. The Tool Use module is responsible for multiple rounds of tool invocation, multimodal data interaction, and tool registration management. RL-Training handles diversified reward computation and tool validation, while WebUI provides a streamlined and user-friendly interactive graphical interface for users.

\subsection{Reinforcement Learning for Tool Invocation Agents}

In the traditional pure-text Chain-of-Thought (CoT) reinforcement learning paradigm, the Markov Decision Process (MDP) defines the "state" by focusing on the model-endogenous information. Specifically, it is composed of the input prompt tokens plus all tokens autonomously generated by the model before the current step. However, when the reinforcement learning scenario is extended to tool invocation, the agent needs to interact deeply with the external environment. A state representation that relies solely on model-endogenous information fails to capture the impact of tool feedback on decision-making.

To address this, we introduce Observation Tokens to reconstruct the MDP state space. These tokens are not generated by the model itself but originate from the results of external tool invocations (such as query results returned by a search interface, execution logs output by a code interpreter, etc.). The observation results are dynamically appended to the trajectory sequence and fed back to the model, serving as input for subsequent decision-making steps. This realizes the closed-loop interaction of "model - tool - environment".

For reinforcement learning with tool invocation, we formally model the MDP state \( s_t \) at step \( t \) as follows:

\[
s_t = \{(X_0, O_0), (X_1, O_1), \ldots, (X_t, O_t)\} = \{X_{\leq t}, O_{\leq t}\}
\]

Where: \( X_{\leq t} = \{X_1, \ldots, X_t\} \) represents the text token sequence, which includes the decision instructions, interaction prompts, or intermediate outputs generated by the model before step \( t \), reflecting the model's endogenous decision trajectory, \( O_{\leq t} = \{O_1, \ldots, O_t\} \) represents the observation token sequence, corresponding to the results returned by tool invocations before step \( t \) (such as search summaries, code execution status, images returned by the environment, etc.), depicting the feedback from the external environment on the model's decision-making.

In state \( s_t \), the model samples and generates an action \( a_t \) (such as a tool invocation instruction, a final answer declaration, etc.) according to the policy \( \pi_\theta(a \mid s_t) \), which serves as the input token for the next interaction step.

The interaction process of agent reinforcement learning for multi-turn tool invocation follows the cyclic logic of "text token generation - tool invocation - observation result feedback". The model generates text tokens to drive tool invocation, and the results returned by the tool are fed back in the form of observation tokens. The two act alternately until the final answer is output or the maximum number of tool invocations is reached. It should be specially noted that, as "environmental feedback", observation tokens are essentially the output of external tools and do not participate in the model loss calculation. This decouples the training dependence between "model decision-making" and "environmental feedback".

\subsection{Key Modules and Processes}
\begin{figure}[h]
    \centering
    \includegraphics[width=1\linewidth]{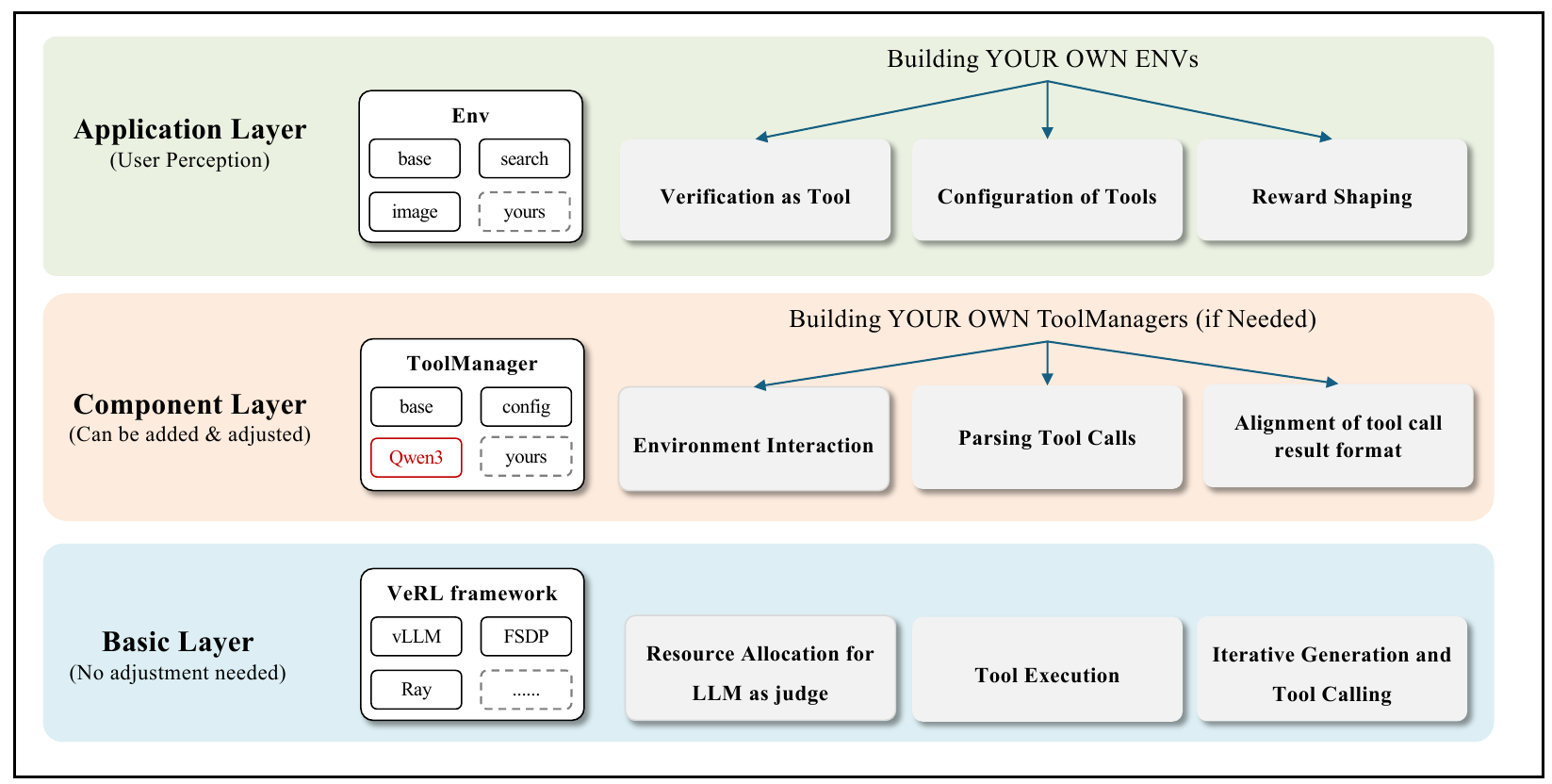}
    \vspace{-6mm}
    \caption{A diagram of the RLFactory layer structure. The basic layer can be used directly without any code modifications. The component layer has a tool manager implemented, and users can also design their tool managers. The application layer provides example search and image environments, allowing users to customize their environments based on specific problems.}
    \label{rlf_layer}
    \vspace{-8pt}
\end{figure}
\subsubsection{Tool Use Environment Construction}
Tool use constitutes the core application scenario for RLFactory to implement agent reinforcement learning. Its primary objective is to build a closed-loop system that enables models to interact with external tools autonomously and efficiently. This process requires addressing two key challenges: the flexible definition of tools and the efficient operation of multi-turn interaction workflows, which together empower models to dynamically invoke tools in complex tasks.

The RLFactory framework supports users in conducting reinforcement learning (RL) training for Multi-Turn Tool Use with minimal code. As illustrated in Figure~\ref{rlf_layer}, we elaborate on the core architecture of the framework across three layers:

\begin{enumerate}
\item \textbf{Application Layer}: Focuses on user task customization. Users need to build a dedicated tool usage environment (Env) and complete tool registration in the MCP format and reward calculation within it. Its core components include:
\begin{itemize}
\item Tool configuration file (\texttt{mcp\_tools.pydata}), which defines the tool call protocol, parameter specifications, etc.
\item Reward calculation logic (\texttt{Env/compute\_score}), which defines how rewards are calculated based on tool interaction results.
\item Result verification tool (\texttt{Env/verify\_tool}), suitable for scenarios where a specific verifier needs to be called based on tool execution results.
\end{itemize}

\item \textbf{Component Layer}: Supports functional expansion and customization. It provides default common components (such as the basic parsing logic of \texttt{Qwen3\_ToolManager}) to meet daily tool interaction needs. For specialized scenarios (e.g., adapting to private tool protocols or custom parsing workflows), users can modify existing components or add new functionality (e.g., building a custom \texttt{ToolManager}), minimizing code changes to accommodate complex tasks.
\begin{itemize}
\item Alignment of tool call result format: Combining tool call results and the original prompt via \texttt{ToolManager/get\_prompt} and post-processing with \texttt{ToolUtils/postprocess\_output} and \texttt{ToolUtils/compose\_final\_output}.
\item Parsing Tool Calls: Parsing tool parameters in model responses via \texttt{ToolManager/parse\_response}.
\item Environment Interaction: Executing a tool environment and returning observations via \texttt{Env/step}.
\end{itemize}

\item \textbf{Foundation Layer}: Ensuring the reuse of core capabilities. It encapsulates the underlying interaction logic used by multiple rounds of tools (e.g., the "generate-parse-call-update" closed-loop workflow), Model Judge infrastructure (including distributed deployment and asynchronous call adaptation), and verl-based native reinforcement learning training mechanisms (e.g., the PPO algorithm and gradient optimization process). Users do not need to worry about the underlying implementation and can directly reuse the stable training and interaction capabilities.
\end{enumerate}

RLFactory abandons narrow definitions of "tools" and adopts a generalized, inclusive design philosophy. Tools are abstracted into three forms, covering diverse interaction needs from single-function programs to complex agents:

\begin{enumerate}
    \item \textbf{Program Tools}: Centered on standardized parameter interaction. Typical examples include:
    \begin{itemize}
        \item Search interfaces: Input query statements, perform network requests and data parsing, and output retrieval results. 
        \item Code interpreters: Receive code text, execute it, and return runtime logs, error messages, or computation results. 
        \item Calculators: Output computation results via numerical operations based on mathematical formulas. 
    \end{itemize}
    These tools rapidly extend models' computational capabilities (e.g., complex mathematical operations) and information-acquisition capabilities (e.g., real-time information retrieval) through direct "input-output" mapping, serving as fundamental components for models to transcend their inherent knowledge and computational boundaries.

    \item \textbf{Model Tools}: Support integration of third-party open-source or closed-source models, leveraging external models' expertise to supplement the trained model's capabilities. For example:
    \begin{itemize}
        \item Invoking GPT-4o can leverage its powerful text comprehension and generation capabilities to add long-text summarization functionality to the model. 
        \item Integrating the Stable Diffusion model enables the model to acquire "text-image" cross-modal generation capabilities. 
    \end{itemize}
    Through standardized model invocation interfaces (e.g., APIs, local model loading), RLFactory enables plug-and-play of model tools without requiring deep modifications to the core framework logic.

    \item \textbf{Agent Tools}: Complex systems combining program modules and model modules, focusing on end-to-end task automation. For example, the "literature research agent" integrates:
    \begin{itemize}
        \item A search tool to acquire literature sources. 
        \item A summarization model tool to extract core viewpoints from literature. 
        \item A citation parsing program to standardize literature citation formats. 
    \end{itemize}
    After inputting a research topic, through multi-turn tool collaboration and result integration, it directly outputs a review report with structured citations. These tools transform multi-step tasks into "one-click" interactions for agents via procedural tool orchestration, significantly improving complex task processing efficiency.
\end{enumerate}

To enable efficient tool management and invocation, RLFactory uses an MCP configuration file (\texttt{mcp\_tools.pydata}) to uniformly define tool metadata. The configuration file includes key information such as tool names, parameter formats (e.g., input parameter types, whether required, default values), and invocation endpoints (e.g., API addresses, local program paths). Users only need to write or modify the configuration file to quickly integrate new tools, without delving into framework code adjustments, truly achieving "low-code" tool expansion.

\subsubsection{Multi-Turn Interaction Flow}
\begin{figure}[h]
    \centering
    \includegraphics[width=1\linewidth]{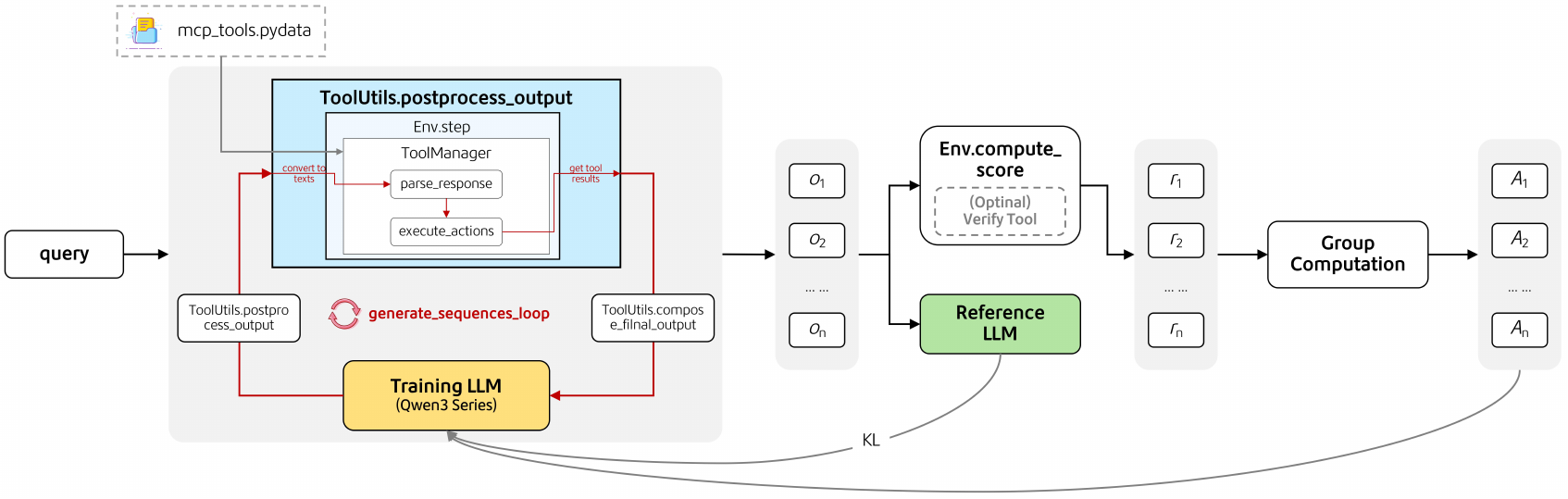}
    \vspace{-6mm}
    \caption{RLFactory multi-round tool call logic diagram using GPRO as an example.}
    \label{rlf_rollout}
    \vspace{-10pt}
\end{figure}
Multi-turn tool invocation is a core part for agents to make dynamic decisions in complex tasks. Figure~\ref{rlf_rollout} shows the multi-turn interaction process using the GRPO algorithm as an example.

RLFactory designs a “Generate - Parse - Invoke - Update” closed-loop interaction flow to ensure continuous and dynamic interaction between the model and tools. Additionally, to avoid the “noise interference” problem caused by model-tool invocations in multi-turn interactions, RLFactory sets a Loss Mask for tool return results.

\begin{enumerate}
    \item \textbf{Generate}: Receive the initial prompt input by the user. The training model generates a response based on the current context (including historical interaction trajectories and tool feedback results). This response is not just a simple text output; it is the model's preliminary decision for the task and may include tool invocation instructions, intermediate reasoning conclusions, or final answer candidates.
    \item \textbf{Parse}: Rely on the ToolManager component to perform structured parsing on the model response. Through predefined tool invocation grammar rules (such as specific function invocation formats and keyword matching), extract tool invocation instructions, including the target tool name and input content (such as the query keywords of a search tool and the execution code snippet of a code interpreter). If no tool invocation intention is identified during parsing, the interaction is determined to be terminated, and the model's reply is directly output as the task result; if a tool invocation is identified, the next interaction link is entered.
    \item \textbf{Invoke}: Based on the Python asyncio asynchronous framework, realize parallel execution of tool invocations. For multi-tool invocation scenarios (such as querying weather and traffic information simultaneously), the asynchronous mechanism allows multiple tool invocation tasks to be executed concurrently, avoiding the situation where a slow response of a single tool (such as a network request timeout) blocks the entire training process and improves the overall efficiency.
    \item \textbf{Update}: Perform formatting processing on the tool return results (such as extracting key information, converting data structures, and filtering redundant content), and append them as observation tokens to the interaction context. The updated context includes the model's historical responses and tool feedback results, providing dynamic feedback for the next-round model invocation. The model can adjust decisions based on the new context, realizing strategy iterative optimization in multi-turn interactions.
\end{enumerate}
\subsection{RL-Training}
\subsubsection{Reward Calculation Strategies}
The core of the reinforcement learning training framework is to design flexible incentive mechanisms and efficient optimization models, enabling the model to learn "when to use tools and how to use tools". RLFactory implements a flexible reward calculation mechanism through the \texttt{envs} module, supporting diverse calculation paradigms to adapt to the characteristics of different tasks. The core includes the following three types of implementations:

\textbf{Reward Based on Rules.}
For tasks with clear success criteria (such as scenarios with verifiable results like NL2SQL, mathematical reasoning), define the reward calculation logic through the \texttt{compute\_score\_with\_rules} method in the environment class. This mechanism, based on the \texttt{verl} framework, ensures a lightweight multi-round tool invocation process. Typical rule dimensions include:
\begin{itemize}
    \item \textbf{Tool Call Format Effectiveness}: Check the formal requirements, such as the integrity of the parameter JSON structure and the existence of necessary fields.
    \item \textbf{Task Completion Degree}: Verify the matching degree between the result and the expected goal (such as the correctness of SQL query results).
    \item \textbf{Efficiency Indicator}: Evaluate the resource consumption to achieve the goal (such as the number of tool calls, interaction rounds).
\end{itemize}

The rule reward function is expressed by a weighted summation of each dimension score:
\begin{equation}
R_{rule}(s, a, s') = \sum_{i = 1}^{n} w_i \cdot r_i(s, a, s')
\end{equation}
Among them: $R_{rule}(s, a, s')$ is the total reward of the $i$-th rule; $(s, a, s')$ respectively represent the current state, model action and post execution state; $r_i$ is the reward component of the $i$-th rule; $w_i$ is the weight of the corresponding rule; $n$ is the total number of rules.

\textbf{Reward Based on Model Evaluation.}
For open-ended tasks (such as scenarios that are difficult to code rules, like knowledge graph search, creative planning), the framework reuses \texttt{verRL reward\_rollout\_wg} dedicated worker nodes and allocates independent resource pools for them. Through \texttt{vlm} to achieve QwQ-32B and other strong model distributed reasoning, efficiently generate reward signals. Workflow includes:
\begin{enumerate}
    \item \textbf{Configuration Activation}: Activate the mechanism through \texttt{reward\_rollout.if\_use\_reward\_rollout=True}.
    \item \textbf{Prompt Construction}: In the \texttt{get\_prompt\_for\_reward} method, generate evaluation prompts based on model output and actual generated content.
    \item \textbf{Distributed Reasoning}: Use pre-deployed resources for parallel evaluation to generate quantitative scores.
    \item \textbf{Score Extraction}: Parse the evaluation results through \texttt{compute\_single\_score\_with\_reward\_rollout\_wg}.
\end{enumerate}

The model evaluation reward function is expressed as:
\begin{equation}
R_{judge}(\tau) = f_{judge}(\tau, c)
\end{equation}
Among them: $R_{judge}(\tau)$ is the model evaluation reward; $\tau$ is the complete interaction trajectory; $c$ is the evaluation criterion (defined through \texttt{prompt}); $f_{judge}(\cdot)$ is the reasoning function of the evaluation model.

\textbf{Tool Verification Reward.}
The \texttt{verify\_tool} method supports using called tools (such as code interpreters, database query tools) to verify model output, especially suitable for tasks that require external operations to evaluate results (such as NL2SQL, code generation). The verification process uses an asynchronous parallel mode to improve efficiency and stores the results in:
\begin{verbatim}
data.non_tensor_batch['reward_model']['ground_truth']['verified_results']
\end{verbatim}
for subsequent scoring.

The tool verification reward function is:
\begin{equation}
R_{verify}(a) = g(T_{verify}(a), y_{expected})
\end{equation}
Among them: $R_{verify}(a)$ is the tool verification reward; $a$ is the action generated by the model (such as SQL statements or code snippets); $T_{verify}(\cdot)$ is the execution function of the verification tool; $y_{expected}$ is the expected result; $g$ is a comparison function that evaluates the matching degree between the actual result and the expectation.

Three types of reward mechanisms can be used independently or in combination. Through the unified interface implementation of the \texttt{Env} class and seamless integration with the training process, it meets the evaluation needs of different task scenarios.

\section{Experiment on Search-R1}
\label{sec:exp}

Search-R1~\cite{jin2025search} is a reinforcement learning framework designed to train LLMs with interleaved reasoning and search capabilities. These language models can learn to perform reasoning and tool calls (e.g., to search engines) in a coordinated manner. Built on veRL, it extends the ideas of DeepSeek-R1 (-Zero) by integrating interleaved search engine access. As an open alternative to OpenAI DeepResearch, it supports research and development of tool-augmented LLM reasoning.

In our reproducibility study of the Search-R1 method, we adopted Qwen3-4B as the base model for reinforcement learning training. The experimental results are shown in Table \ref{tab:search_r1_performance}. The trained model achieved a notable performance of 0.486 on the evaluation metric, outperforming the larger-parameter model Qwen2.5-7B-Instruct-GRPO (0.429) trained with the same Search-R1 technique. Notably, despite the limited computational resources, our method demonstrated exceptional computational efficiency, with the training throughput improved by 6.8 times, highlighting its great potential in optimizing resource utilization.

\begin{table}[htbp]
    \centering
    \caption{Performance comparison of different Search-R1 variants on NQ dataset.}
    \label{tab:search_r1_performance}
    \begin{tabular}{lccc}
        \toprule
        Method & Test Score (NQ) & Convergence Time & Resources \\
        \midrule
        Search-R1-Qwen2.5-3B-Instruct-GRPO & 0.421 & 23h & A100*8 \\
        Search-R1-Qwen2.5-7B-Instruct-GRPO & 0.473 & 36h & A100*8 \\
        Search-RL-Qwen3-4B-Instruct-GRPO & 0.486 & 5h & A100*8 \\
        \bottomrule
    \end{tabular}
    \vspace{-0.5em}
\end{table}

\begin{figure}[htbp]
    \centering
    \includegraphics[width=0.8\linewidth]{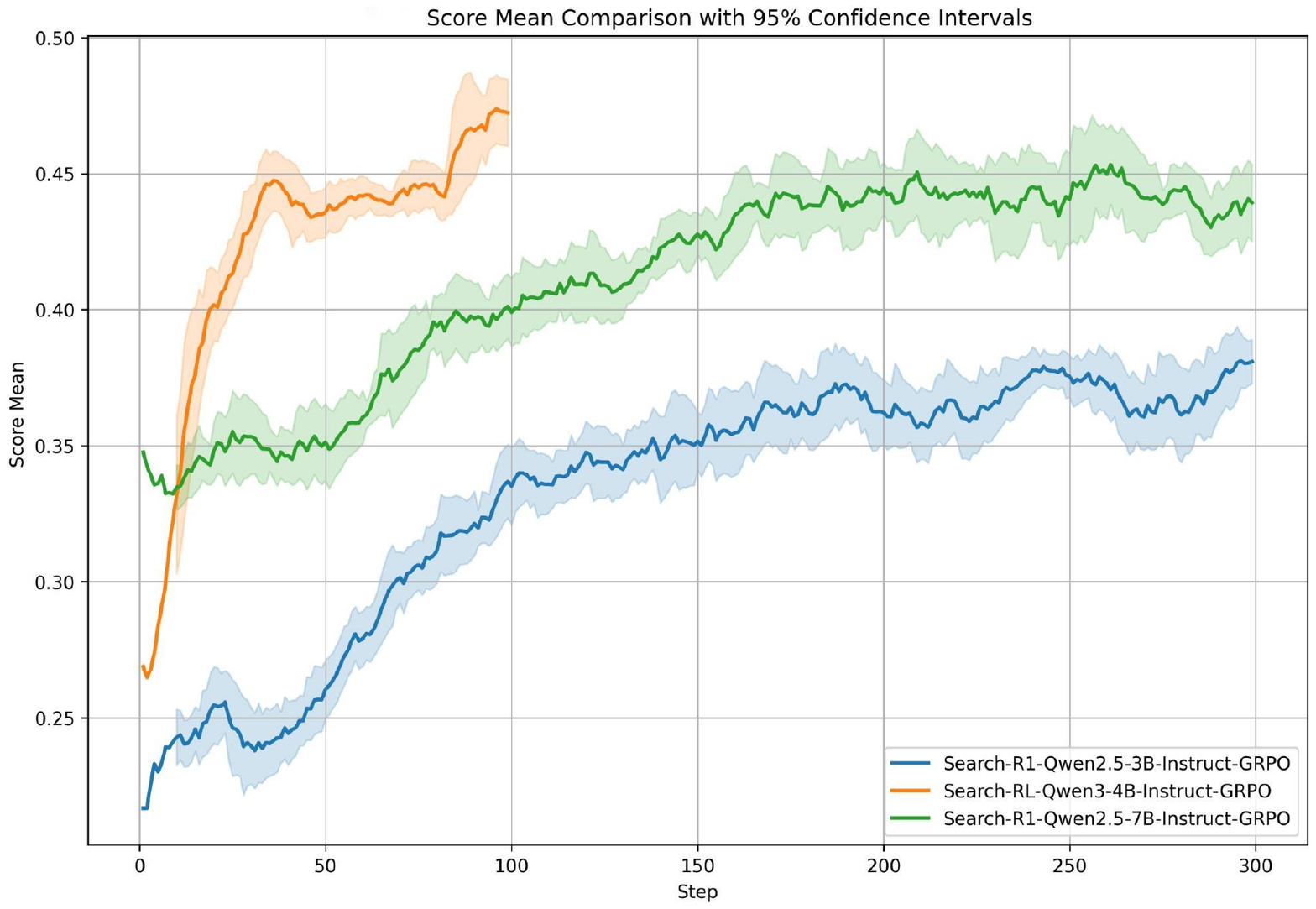}
    \vspace{-6mm}
    \caption{Mean reward score trends across different base model.}
    \label{fig:Search-R1-critic-score}
    \vspace{-10pt}
\end{figure}

Figure \ref{fig:Search-R1-critic-score} visualizes the dynamic changes in the mean Critic scores during the training process of different Search-R1 variants, where the shaded areas represent the 95\% confidence intervals. The curves illustrate the convergence trends: Search-RL-Qwen3-4B-Instruct-GRPO (green), trained based on the RLFactory framework, shows stable improvement and more competitive performance, which is consistent with the quantitative results in Table \ref{tab:search_r1_performance} and reflects the training stability and high efficiency of RLFactory.

\section{Conclusion}
\label{sec:con}

In this paper, we propose RLFactory, a plug-and-play reinforcement learning post-training framework designed to enhance the capabilities of large models (LMs) in multi-turn tool usage scenarios. RLFactory addresses core challenges in tool invocation stability, adaptability, and reward computation diversity. By introducing observation labels to reconstruct the MDP state space, RLFactory facilitates an efficient model-tool-environment feedback loop, while its generate-parse-invoke-update workflow ensures robust multi-turn interactions. Experimental results on the Search-R1 benchmark validate the effectiveness and efficiency of RLFactory. While RLFactory demonstrates potential in enhancing multi-turn tool usage, future work will focus on expanding its compatibility with more types of environments and complex task scenarios. We also aim to improve the reward mechanism to better handle highly open-ended tasks and support a wider range of LLMs, thereby achieving broader applicability. In summary, RLFactory provides a low-barrier, adaptable solution for advancing LLM agents in practical tool augmentation scenarios, paving the way for more robust and efficient model-tool collaboration.

\newpage
\section*{Authors}
Authors are listed alphabetically by surname within each role.

\subsection*{Project Lead}
\begin{itemize}
  \item \href{mailto:chaijiajun@meituan.com}{Jiajun Chai}
  \item \href{mailto:yinguojun02@meituan.com}{Guojun Yin}
\end{itemize}

\subsection*{Core Contributors}
\begin{itemize}
  \item Chengqi Dong
  \item Hang He
  \item Yi Jia
  \item Jiwen Jiang
  \item Xiaoguang Li
  \item Xiaohan Wang
  \item Siyu Xia
  \item Zekun Xu
  \item Chuhuai Yue
\end{itemize}

\subsection*{Supervision}
\begin{itemize}
  \item Wei Lin
\end{itemize}

\newpage
\medskip
{
    \small
    \bibliographystyle{unsrt}
    \bibliography{main}
}

\newpage
\appendix



\end{document}